%% file: main.tex
\documentclass[10pt,twocolumn,letterpaper]{article}
\input{preamble}
\usepackage{cvpr}              % To produce the CAMERA-READY version
\usepackage{graphicx}
\usepackage{amsmath}
\usepackage{amssymb}
\usepackage{booktabs}
\usepackage[pagebackref,breaklinks,colorlinks]{hyperref}

\usepackage[capitalize]{cleveref}
\crefname{section}{Sec.}{Secs.}
\Crefname{section}{Section}{Sections}
\Crefname{table}{Table}{Tables}
\crefname{table}{Tab.}{Tabs.}

\def\cvprPaperID{5} 
\def\confName{ICCV}
\def\confYear{2025}
\begin{document}
\title{UniPaint: Unified Space-time Video Inpainting via Mixture-of-Experts}

\author{Zhen Wan\affmark[1]\quad
        Chenyang Qi\affmark[2] \quad
        Zhiheng Liu\affmark[3] \quad
        Tao Gui\affmark[1] \quad
        Yue Ma\affmark[2]$^\ast$\\
        {\affmark[1]Fudan University\quad\quad\affmark[2] HKUST \quad\quad\affmark[3] HKU}
        }

\input{figs/0_teaser}
\footnotetext{$^\ast$Corresponding author. E-mail: \texttt{mayuefighting@gmail.com}}
%%%%%%%%% ABSTRACT
\input{sec/0_abstract}

%%%%%%%%% BODY TEXT

\input{sec/1_intro}
\input{figs/1_0_overview}
\input{sec/2_related}
\input{figs/2_qualitative}
\input{sec/3_method}

\input{sec/4_experiments}

\input{sec/5_conclusion}
\input{sec/X_supp}

%%%%%%%%% REFERENCES
{\small
\bibliographystyle{ieee_fullname}
\bibliography{main.bib}
}

\end{document}

%% file: preamble.tex
\usepackage[dvipsnames]{xcolor}
\def\confName{ICCV}
\def\confYear{2025}

\definecolor{myred}{RGB}{220,50,47} %
\definecolor{mygreen}{RGB}{133,153,0}
\definecolor{commentcolor}{RGB}{133,153,0}

\definecolor{mygreen}{rgb}{0.29, 0.7, 0.48}

\newcommand*{\affmark}[1][*]{\textsuperscript{#1}}
 

%% file: figs/0_teaser.tex
\twocolumn[{
\renewcommand\twocolumn[1][]{#1}
\maketitle
\begin{center}
    \centering
    \vspace*{-.5em}
    \includegraphics[width=1\linewidth]{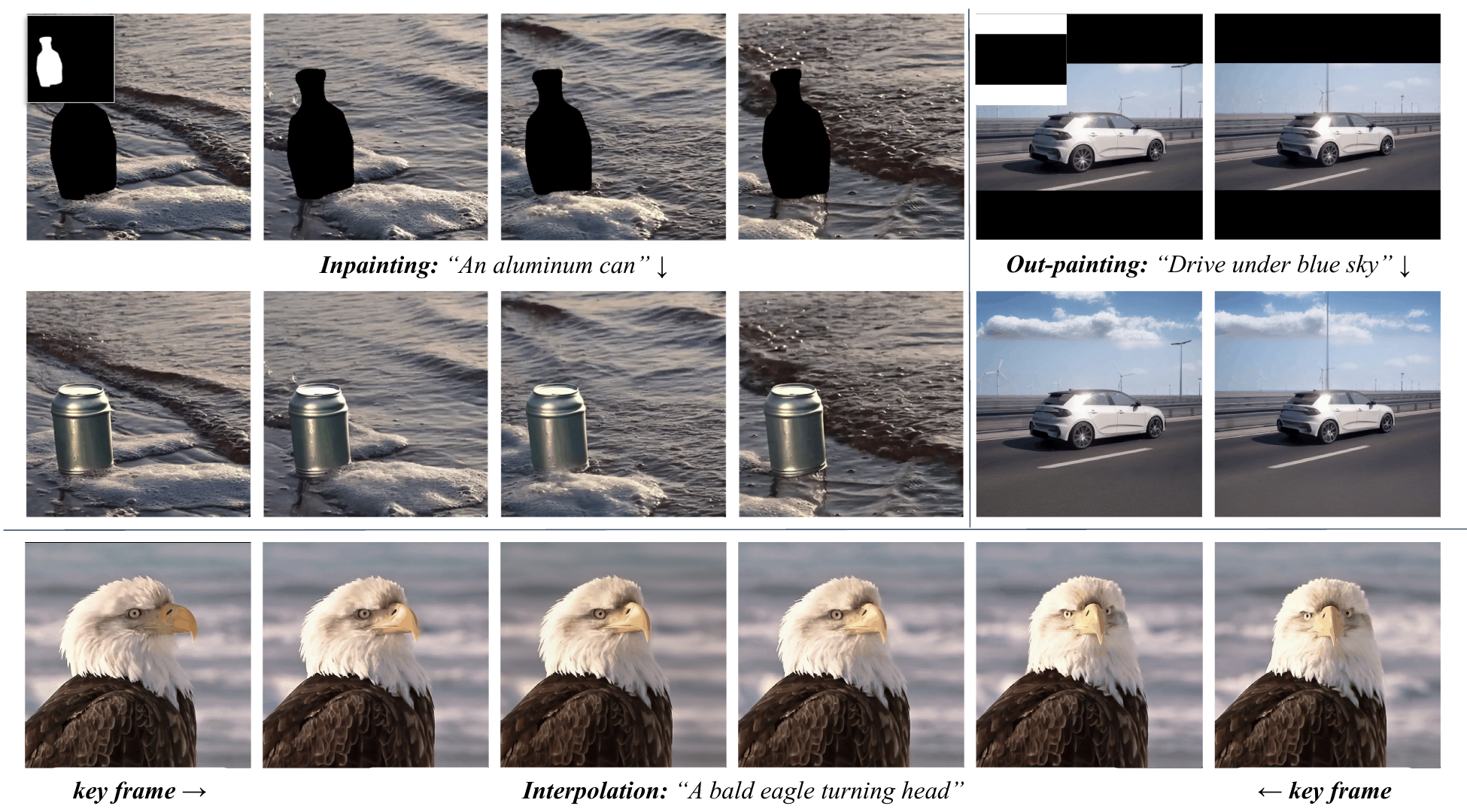}
    \captionof{figure}{
    \textbf{The visual results of unified space-time video inpainting.} 
    We introduce a space-time video inpainting method that is versatile across a spectrum of tasks. Displayed frames are uniformly selected from videos of different space-time inpainting scenarios. For inpainting and outpainting, the first row in the figure contains the source videos and the target regions, while the bottom row shows the results. For interpolation, two keyframes and generated interpolations in between are displayed.
    }
    \label{fig: teaser}
\end{center}
}]

%% file: sec/0_abstract.tex
\newcommand{\st}{spatial-temporal }

\begin{abstract}

In this paper, we present \textbf{UniPaint}, a unified generative space-time video inpainting framework that enables spatial-temporal inpainting and interpolation.
Different from existing methods that treat video inpainting and video interpolation as two distinct tasks, 
we leverage a unified inpainting framework to tackle them and observe that these two tasks 
can mutually enhance synthesis performance. 
Specifically, we first introduce a plug-and-play space-time video inpainting adapter, which can be employed in various personalized models. The key insight is to propose a Mixture of Experts (MoE) attention to cover various tasks. Then, we design a \st masking strategy during the training stage to mutually enhance each other and improve performance.
UniPaint produces high-quality and aesthetically pleasing results, achieving the best quantitative results across various tasks and scale setups. The code and checkpoints are available at \url{https://github.com/mmmmm-w/UniPaint}.

\end{abstract}

%% file: sec/1_intro.tex
\section{Introduction}
\label{sec:intro}
Video inpainting aims to restore missing spatial and temporal regions in a source video while preserving visual coherence and temporal consistency. As a foundational task in computer vision, video inpainting has attracted considerable academic exploration in ~\cite{zi2024cococoimprovingtextguidedvideo, zhang2024avidanylengthvideoinpainting}. This technology has widespread applications in fields of the film industry, automatic advertising, and content creation on social media platforms, \textit{etc}. 

Recently, diffusion model~\cite{ho2020denoising, song2020denoising, song2022diffusion} has emerged as the mainstream approach for image inpainting~\cite{paintbyword_andonian2021paint, controlnet_zhang2023adding, smartbrush_xie2023smartbrush, ju2024brushnetplugandplayimageinpainting}, demonstrating realistic and contextually consistent results. Imagenator~\cite{imageneditor_wang2023imagen} leverages the pre-trained text-to-image diffusion model~\cite{imagen_saharia2022photorealistic} to modify the source image. BrushNet~\cite{ju2024brushnetplugandplayimageinpainting} and Powerpaint~\cite{zhuang2023taskPowerPaint} propose a plug-and-play approach to improve the inpainting precision and generation quality. Unlike single-frame image inpainting, maintaining temporal consistency is also crucial in video inpainting. VideoComposer~\cite{wang2023videocomposer} applies mask-based constraints across frames, while CoCoCo~\cite{zi2024cococoimprovingtextguidedvideo} and AVID~\cite{zhang2024avidanylengthvideoinpainting} employ the structure guidance and frame-by-frame masking strategy, enabling flexible user control with semantics. 

Despite these advancements, existing approaches primarily focus on the spatial dimension, leaving an important question unanswered: \textbf{\textit{Can a unified framework effectively address both spatial and temporal inpainting?}}

To address this challenge, we present the \textbf{UniPaint}, the first unified diffusion-based framework for space-time video inpainting. A comparison is provided in \cref{tab:comparison}.
Different from existing methods that consider video interpolation as a separate task, 
we treat video interpolation as an inpainting problem that spans both the temporal and spatial dimensions. Then, we integrate them into a unified space-time inpainting task (\cref{fig: teaser}). Specifically, to preserve the generative capabilities of the pretrained model, we introduce a plug-and-play space-time video inpainting adapter rather than optimizing all the parameters of the foundation model. To cover various tasks, we propose the Mixture of Experts (MoE) attention (\cref{fig: attn_change}), which leverages different experts to handle various tasks. As shown in \cref{tab: quan_spatial,tab: quan_temporal}, our experiments reveal that our integrated approach mutually enhances both spatial and temporal inpainting performance. Additionally, during the training stage, we design a spatial-temporal masking strategy to facilitate the space-time video inpainting. We perform extensive quantitative and quality experiments. The comprehensive results demonstrate that UniPaint excels across a range of video inpainting tasks,  achieving state-of-the-art performance in space-time video inpainting.

\input{figs/1_3_attn_change}
Our contributions can be summarized as follows:
\begin{itemize}
    \item  We present a novel insight that integrates video inpainting and interpolation into a unified space-time video inpainting framework, proposing UniPaint, the first unified diffusion-based framework to tackle space-time video inpainting. 

    \item  To achieve robust space-time video inpainting, we first introduce a plug-and-play space-time video inpainting adapter to facilitate powerful generative ability.  Then, the Mixture of Experts (MoE) attention and spatial-temporal masking strategy are designed to handle task diversity and enhance performance. 
    
    \item We conduct extensive quantitative and qualitative evaluations, including video inpainting, video outpainting, and video interpolation. The experiment results show the superiority of the proposed method.

\end{itemize}

%% file: figs/1_3_attn_change.tex
\begin{figure}
    \centering
    \vspace{-1em}
    \includegraphics[width=0.8\linewidth]{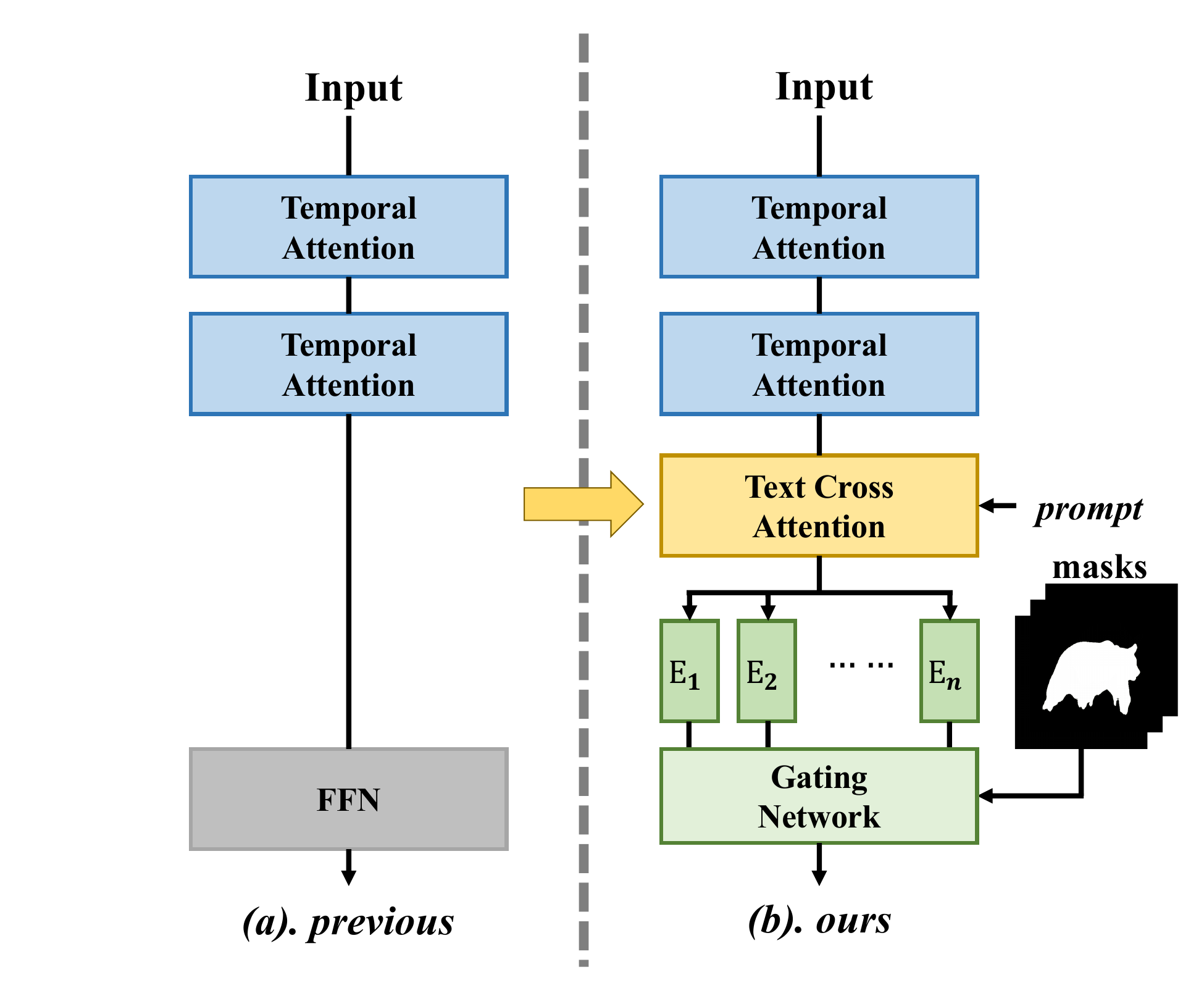}
    \vspace{-1.em}
    \caption{\textbf{Previous attention \textit{v.s.} MoE attention.} We incorporate our Space-time Inpainting Adapter with MoE attention, providing better adaptability and textual alignment.
    }
    \label{fig: attn_change}
    \vspace{-.5em}
\end{figure}

%% file: figs/1_0_overview.tex
\begin{figure*}
\setlength{\linewidth}{\textwidth}
\setlength{\hsize}{\textwidth}
\centering
\includegraphics[width=1\linewidth]{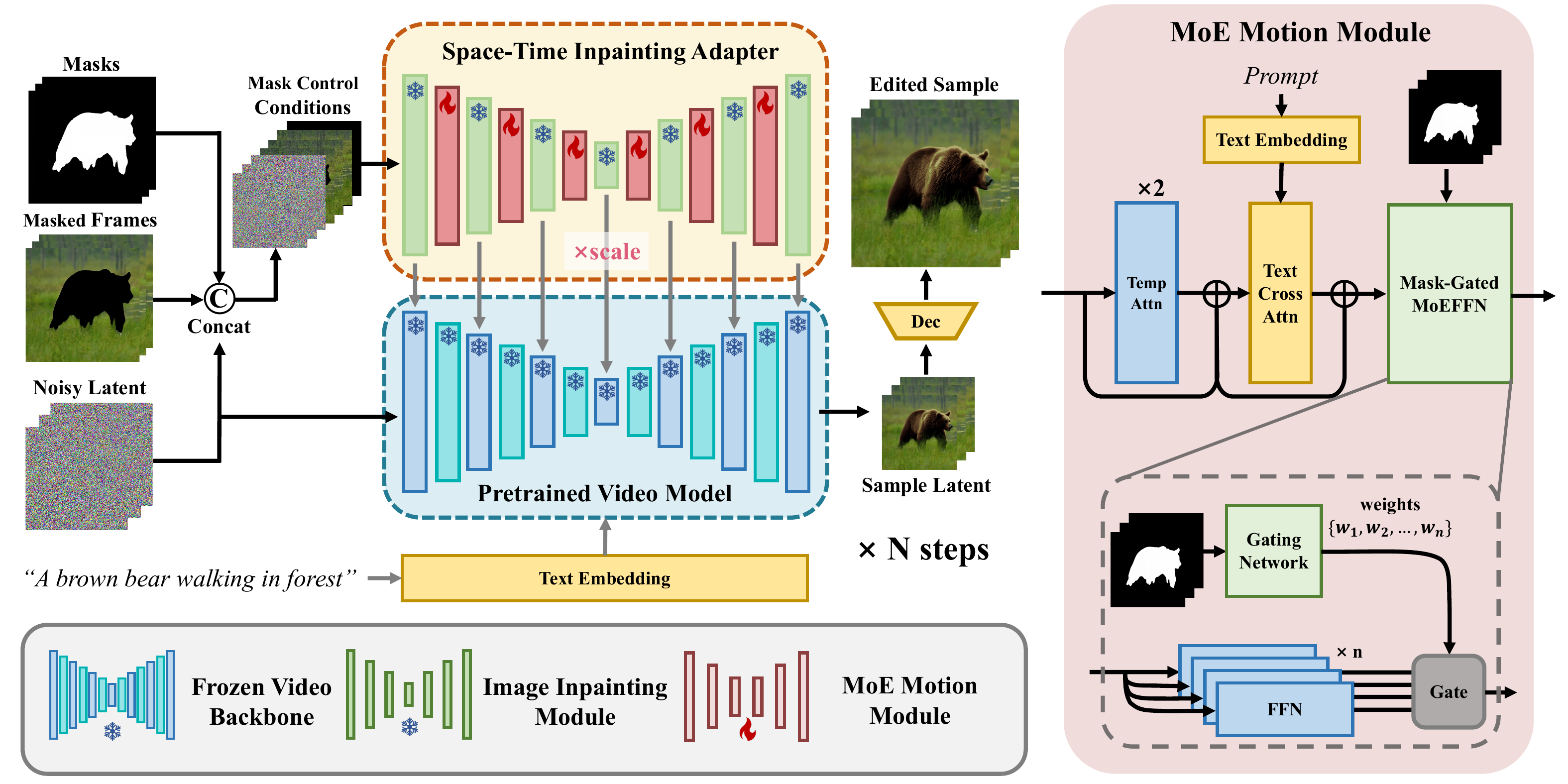}
\caption{\textbf{Overview of our method.}
As shown in the figure, \emph{UniPaint} outputs an inpainted video given the mask and masked video input. The noise, masked frames, and masks are concatenated as the input to the Space-Time Inpainting Adapter. The feature extracted by the adapter is added to the pretrained video model with a custom scale. The mask is also input to the gating network of the MoE Motion Module.
}
\label{fig: overview}

\vspace{-1em}
\end{figure*}

%% file: sec/2_related.tex
\section{Related Work}

\paragraph{Video generation} has obtained significant attention from both the public and academic circles. Recent advancements in video generation have leveraged diffusion models to achieve impressive visual quality~\cite{imagenvideo_ho2022imagen,gen1_esser2023structure, gen2, pikalab, soravideoworldsimulators2024, stablevideodiffusion_blattmann2023stable, videopoet_kondratyuk2023videopoet, videocrafter2_chen2024videocrafter2, Dynamicrafter_xing2023dynamicrafter, 
tuneavideo_wu2023tune, khachatryan2023text2video, animatediff_guo2023animatediff,chen2023controlavideo,yin2023dragnuwa,modelscope_wang2023modelscope,fan2023hierarchical,zhang2023show1,guo2023sparsectrl,jiang2023videobooth,ma2024magicme,xie2023dreaminpainter,ma2024followyouremoji,ma2025followyourclick,ma2022visual,ma2023magicstick,ma2022simvtp,ma2024follow,ma2025followcreation,ma2025followyourmotion,yan2025eedit, zhang2025magiccolor,zhu2024instantswap, wang2024cove, feng2025dit4edit, chen2024follow}, marked by both closed-source and open-source models. Closed-source models like Sora~\cite{soravideoworldsimulators2024}, Pika~\cite{pikalab}, VideoPoet~\cite{videopoet_kondratyuk2023videopoet}, Gen1\cite{gen1_esser2023structure}, Gen2~\cite{gen2} and Kling~\cite{kling} offer high-resolution, long-duration videos. However, their proprietary methodologies and datasets limit research access and reproducibility. In contrast, open-source methods have accelerated innovation in research communities by providing accessible frameworks. For example, Tune-A-Video~\cite{tuneavideo_wu2023tune} minimizes tunable parameters requirement during the adaptation stage for zero-shot video generation while Text2Video-Zero~\cite{khachatryan2023text2video} employs training-free latent code manipulation to produce videos without extensive training. AnimateDiff~\cite{animatediff_guo2023animatediff} keeps image modules static, training only motion components, which allows integration with customized T2I models. Similarly, VideoCrafter~\cite{videocrafter2_chen2024videocrafter2} introduces temporal motion layers for high-quality text-to-video generation, while DynamiCrafter~\cite{Dynamicrafter_xing2023dynamicrafter}
further uses keyframes as guidance for temporal extension and interpolation. Stable Video Diffusion~\cite{stablevideodiffusion_blattmann2023stable} utilizes well-curated datasets with refined captioning, and ModelScope~\cite{modelscope_wang2023modelscope} incorporates spatial-temporal blocks to ensure frame consistency. Together, these advancements underscore the robustness and versatility of diffusion-based models in the video generation landscape.

\input{tab/0_compare}

\paragraph{Text-guided image inpainting} has also seen significant improvements through the application of diffusion models. Recent methods in text-guided image inpainting leverage diffusion models to achieve realistic and contextually consistent results~\cite{paintbyword_andonian2021paint, avrahami2022blended, cogview2_ding2022cogview2, Diffedit_couairon2022diffedit, imageneditor_wang2023imagen, ju2024brushnetplugandplayimageinpainting}. Latent Blended Diffusion~\cite{avrahami2022blendedldm} integrates generated and original image features, balancing foreground and background elements through a blending approach in latent space. Techniques like Imagenator~\cite{imageneditor_wang2023imagen} and Diffusion-based Inpainting~\cite{rombach2022high} adapt pre-trained text-to-image models to handle masked inputs, allowing for precise control over edited areas. Additionally, Brushnet~\cite{ju2024brushnetplugandplayimageinpainting} introduces a mask-conditioned control branch trained on object-centric datasets, enabling highly localized inpainting adjustments. These diffusion-based approaches deliver refined inpainting outcomes that align closely with both the input context and text prompts.

\paragraph{Video inpainting} extends the capabilities of image inpainting to the temporal domain, requiring models to maintain consistency across frames while filling in missing or occluded content. Recent works incorporate pre-trained image models for temporally coherent inpainting~\cite{wang2023videocomposer, zhang2024avidanylengthvideoinpainting, zi2024cococoimprovingtextguidedvideo}. Some leverage pre-trained image models for video inpainting, such as using DDIM~\cite{song2020denoising} inversion to ensure consistent latent representations~\cite{pix2video_ceylan2023pix2video, tokenflow_geyer2023tokenflow, fatezero_qi2023fatezero, editavideo_shin2023edit, wang2023zero, tuneavideo_wu2023tune}. VideoComposer~\cite{wang2023videocomposer} employs mask-based constraints across frames for targeted inpainting, though it may lack flexibility due to its uniform masking approach. Advanced models like AVID~\cite{zhang2024avidanylengthvideoinpainting} and CoCoCo~\cite{zi2024cococoimprovingtextguidedvideo} dynamically adjust the masked regions frame-by-frame, achieving more precise control. However, they face challenges in generalizing to broader tasks like outpainting or interpolation. These advancements illustrate the progress and challenges of achieving seamless, text-guided video inpainting that preserves temporal consistency.

\paragraph{Video frame interpolation (VFI)}, or temporal inpainting in our context, is also a well-established problem in computer vision that has been extensively tackled in recent literature~\cite{vfisurvey}. Some of the most recent methods employ diffusion models to improve VFI by introducing probabilistic frameworks that address the ambiguity of large, nonlinear motion patterns~\cite{Danier_2024, jain2024videointerpolationdiffusionmodels, voleti2022mcvd}. LDMVFI~\cite{Danier_2024} and MCVD~\cite{voleti2022mcvd} utilize diffusion-based approaches, generating frames with enhanced coherence in complex scenes. VIDIM\cite{jain2024videointerpolationdiffusionmodels}, another recent diffusion-based method, differs by operating directly in pixel space and generating full video sequences for superior motion quality. While conventional benchmarks\cite{baker2011database,butler2012naturalistic,pont2017davis,soomro2012ucf101,xue2019video} often assume mostly linear motion, diffusion-based VFI models like VIDIM demonstrate robustness in cases of significant temporal gaps or complex motion, formulating VFI as a generative problem rather than merely pixel correspondence problem.

%% file: tab/0_compare.tex
\begingroup
\setlength{\tabcolsep}{0pt} 
\renewcommand{\arraystretch}{1.0} 
\vspace{-.5em}
\begin{table*}[!htb]\centering

\begin{tabular*}{0.8\linewidth}{@{\extracolsep{\fill}} lcccc}
\toprule
Model & Plug-and-Play& Spatial Inpainting & Temporal Inpainting & Shape-Aware \\ \hline
VideoComposer\cite{wang2023videocomposer} & \checkmark & \checkmark &  &  \\
AVID\cite{zhang2024avidanylengthvideoinpainting} &  & \checkmark &  &\checkmark  \\
CoCoCo\cite{zi2024cococoimprovingtextguidedvideo} &  & \checkmark & &\checkmark  \\
VIDIM\cite{jain2024videointerpolationdiffusionmodels} &  &  & \checkmark &  \\ \hline
\emph{UniPaint (Ours)} & \checkmark & \checkmark & \checkmark & \checkmark \\ 
\bottomrule
\end{tabular*}
\captionsetup{width=.8\textwidth}
\caption{\textbf{Comparison of \emph{UniPaint} with previous video inpainting methods.} \emph{UniPaint} offers the advantage of being plug-and-play with pretrained video model. Moreover, it allows for flexible control over the scale of inpainting and is designed to be aware of both the mask shape and the unmasked content.}
\vspace{-.5em}
\label{tab:comparison}
\end{table*}
\endgroup

%% file: figs/2_qualitative.tex
\begin{figure*}[ht]
\setlength{\linewidth}{\textwidth}
\setlength{\hsize}{\textwidth}
\centering
\includegraphics[width=1\linewidth]{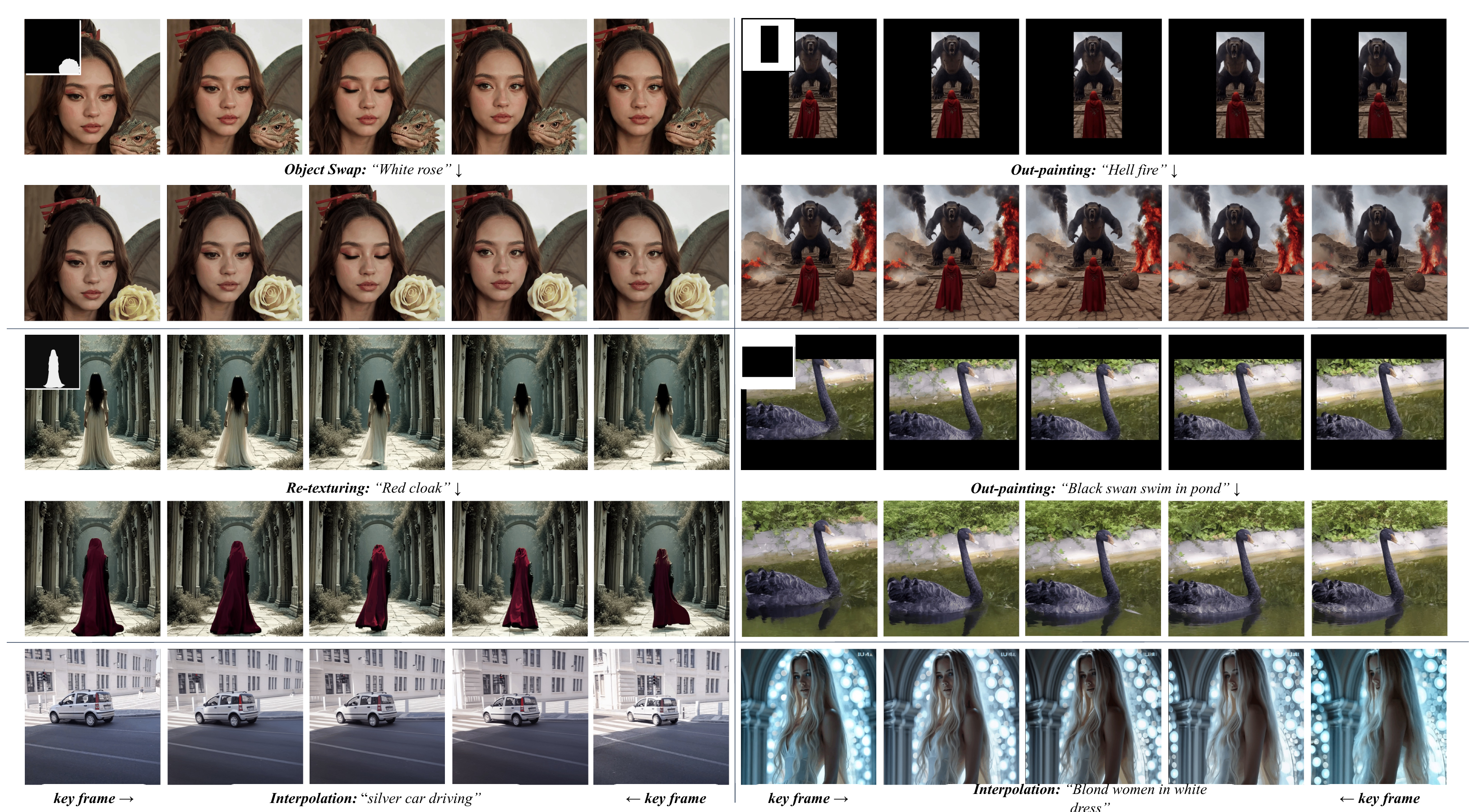}
\vspace{-2em}
\caption{\textbf{Inpainting on videos of different cases.} We employ our method on various scenarios of inpainting. Our method can be applied to both spatial and temporal inpainting cases with arbitrary mask shapes. The caption in the middle represents the inpainting type and prompt guidance for each video. 
}
\label{fig: qual}
\vspace{-1.em}
\end{figure*}

%% file: sec/3_method.tex
\section{UniPaint}
UniPaint is a unified generative \st video inpainting framework capable of both spatial and temporal inpainting (interpolation). While previous works treat spatial and temporal inpainting as distinct tasks\cite{zhang2024avidanylengthvideoinpainting,zi2024cococoimprovingtextguidedvideo, jain2024videointerpolationdiffusionmodels, Danier_2024}, our experiments demonstrate they can be unified under the same mask-filling framework. Different tasks correspond to different types of masks, as shown in \cref{fig: mask}. Given a source video, a \st mask sequence and a text prompt, our objective is to fill in the indicated region following the text guidance, while keeping the out-of-mask video portion consistent.

\input{figs/1_1_mask}

Under this unified framework, we introduce our method, UniPaint, as shown in \cref{fig: overview}. Our approach is built on top of a diffusion-based text-guided video generation model\cite{animatediff_guo2023animatediff}, then adapts it to \st video inpainting model with our Space-time Inpainting Adapter, a plug-and-play mask-conditioned control branch for pixel-level alignment in unmasked area. We further enhance the model's flexibility with MoE attention to actively adapt to different mask types. Moreover, we introduce a novel training procedure that includes both spatial and temporal inpainting cases to boost the model's synthetic ability.

\subsection{Preliminaries}

The diffusion model is defined to approximate the probability density of training data by reversing the Markovian Gaussian diffusion processes~\cite{ho2020denoising, song2020denoising}.
Consider an input video $x_0$, we conduct a forward Markov process described as:

\begin{equation}
\label{equ: diffuse}
q\left(x_t \mid x_{t-1}\right)=\mathcal{N}\left(\sqrt{1-\beta_t} x_{t-1}, \beta_t \mathrm{I}\right),
\end{equation}

where $t = 1, \ldots, T$ indicates the number of diffusion steps, with $\beta_t$ controlling the noise level at each step.
A neural network $\epsilon_\theta$ learns to reverse this process, approximating noise $\epsilon_t$ to restore $x_{t-1}$ from $x_t$ using the relation $x_{t-1} = \frac{1}{\sqrt{\alpha_t}}\left(x_t-\frac{1-\alpha_t}{\sqrt{1-\bar{\alpha}_t} } \epsilon_t \right)$, with $\alpha_t = 1 - \beta_t$ and $\bar{\alpha}_t=\prod_{i=1}^t \alpha_i$, as per~\cite{ho2020denoising}.
For conditional diffusion, in our case, text-guided inpainting, we introduce conditions into $\epsilon_\theta$ without altering the process.
Our training objective can be formulated as:

\begin{equation}
    \label{equ: img loss}
    \mathcal{L} = \mathbb{E}_{\epsilon \sim \mathcal{N}(0, I)}\left[\left\|\epsilon-\epsilon_\theta\left(x_t, t, \mathbf{c}\right)\right\|_2^2\right],
\end{equation}

where $\mathbf{c}$ denotes the conditional inputs.
In our case, $\mathbf{c}=\left(x_m, m, \tau_\theta(y)\right)$, where $m$ is a binary mask indicating the region to modify, $x_m=x_0 \odot (1 - m)$ is the region to preserve, $y$ represents the corresponding textual description while $\tau_\theta \left( \cdot \right)$ embodies a text encoder that transposes the string into a sequence of vectors. Classifier-free guidance~\cite{ho2022classifier} and efficient sampling approaches such as DDIM~\cite{song2020denoising} or PNDM~\cite{liu2022pseudo} can be applied during inference.

\subsection{Space-time inpainting adapter}

The integration of masked features into the pre-trained diffusion network is handled through an additional branch that decouples feature extraction of masked frames from the main video generation process. In previous approaches, mask conditions were concatenated directly with the noisy latent inputs in the main branch \cite{zhang2024avidanylengthvideoinpainting, zi2024cococoimprovingtextguidedvideo}. While effective, this method limits flexibility, as it requires modifications to the model backbone due to inflated input dimensions. Inspired by recent image inpainting works \cite{controlnet_zhang2023adding, ju2024brushnetplugandplayimageinpainting}, we employ a dual-branch architecture that maintains model modularity and flexibility. 

In our setup, the additional branch takes as input the noisy latent, masked frame latent, and downsampled mask, which are concatenated together (see \cref{fig: overview}). The noisy latent provides generative information, guiding the inpainting process to maintain semantic coherence. The masked frame latent, extracted using a Variational Autoencoder (VAE) \cite{vae_kingma2013auto}, is aligned with the data distribution of the pretrained UNet \cite{unet_ronneberger2015u}. The mask is resized to match the latent dimensions via cubic interpolation, ensuring consistent input scaling. UniPaint utilizes a pretrained inpainting control model \cite{ju2024brushnetplugandplayimageinpainting} for feature extraction, with an option to leverage convolutional layers from pretrained text-to-image (T2I) models. This enhances UniPaint’s adaptability and leverages the diffusion model’s pretrained weights for robust feature extraction. The feature insertion operation is formulated as: 
\begin{equation}
\label{eq: insertion}
    \epsilon_{\theta}(\cdot)_{i} = \epsilon_{\theta}(\cdot)_{i} +
    \omega_s \cdot \epsilon_{\theta}^{A}([z_{t}, z_0^{m}, m^{resized}], \tau_\theta(y),t)_i
\end{equation}
where $\epsilon_{\theta}(\cdot)_{i}$ indicates the feature of the i-th layer in main branch $\epsilon_{\theta}$ with $i \in [1, n]$, where $n$ is the number of layers. The same notation applies to $\epsilon_{\theta}^{A}$ which denotes our Space-time Inpainitng Adapter. $\epsilon_{\theta}^{A}$ takes the concatenation of the present noisy latent $z_{t}$, the masked frame latent $z_0^{m}$ and the resized masks $m^{resized}$ as input, with the concatenation operation denoted as [·]. The adapter also accepts text guidance with $\tau_\theta \left( \cdot \right)$. $\omega_s$ is the preservation scale used to adjust the influence of the adapter on pretrained diffusion model.

\subsection{Mixture of Experts Attention}

The diversity of editing scenarios in video inpainting can be generalized by the variation in mask shapes. For instance, in inpainting tasks, the mask typically covers a small, localized area within each frame, while in outpainting tasks, it occupies the marginal regions of the frames. Interpolation, on the other hand, can be formulated as a temporal masking task, where the frames between keyframes are masked. Each of these scenarios requires the motion module to focus on different aspects of spatial and temporal information. To enable adaptive behavior across diverse editing cases, we equip our motion modules with Mixture of Experts (MoE) attention mechanism, as illustrated in \cref{fig: overview,fig: attn_change}.

Our MoE attention module includes two temporal attention layers, a textual cross-attention layer, and a set of expert feedforward networks (FFNs), represented as $\mathbf{E} = {e_\theta^1, …, e_\theta^n}$, where each $e_\theta^i$ serves as an expert. A gating function, $\epsilon_{\theta}^{G}$, takes the resized mask  $m^{\text{resized}}$  as input and determines the weight vector $\mathbf{W} = [w_1, …, w_n], \sum_{i=1}^{n} w_i = 1$, for each expert. The output of the MoE attention is a weighted sum of the outputs from all experts. To extract the shape information from the mask, we apply multiple 3D downsampling convolution layers followed by adaptive average pooling. This output is then passed through a linear layer to project it into the weight space for the experts. Formally, the MoE attention can be expressed as:
\begin{align}
\label{eq: moeffn}
\mathbf{W} &= \epsilon_{\theta}^{G}(m^{\text{resized}}),\ \mathbf{W} \in \mathbb{R}^{n \times 1}, \\
z^{\prime} &= \sum_{i=1}^{n} w_i \times e_i(z),
\end{align}
where $z$ denotes the input to the MoE layer and  $z^{\prime}$  represents the output after weighting.

% To further enhance alignment with the textual prompt, we incorporate an additional textual cross-attention layer. This layer integrates motion-related information derived from the text prompt to guide the model’s inpainting decisions. To optimize memory efficiency, we flatten the visual input  $z \in \mathbb{R}^{f \times w_1 \times h_1}$  into a vector and use it as the query, while the text embeddings serve as the key and value. This approach reduces computational overhead while preserving essential text-video alignment for detailed and coherent inpainting.

\subsection{Space-time Mask Training}
\label{sec: mix_mask_training}
\input{figs/1_2_mask_generation}
Ensuring mask consistency across frames is essential for text-video alignment. Training a video diffusion model on random masks can destabilize training and reduce inpainting accuracy, limiting the model’s ability to learn motion information relative to the given prompt. Recent advances in video segmentation have facilitated segmenting videos with text prompts. To this end, we introduce a segmentation-based mask generation process for training shown in \cref{fig: mask_generaiton}. We employ GroundingDINO~\cite{groundingdino_liu2023grounding} to annotate the first frame in each training video. Specifically, we detect the first frame and retrieve phrases with associated bounding boxes. Using the bounding box from the initial frame as input, SAM2~\cite{ravi2024sam2segmentimages} propagates the segmentation through subsequent frames, generating object segmentations that correspond to the text prompt. This approach enables the creation of text-mask pairs for each video clip.

To enhance adaptability across various inpainting scenarios, we use a mixed mask training strategy that combines four mask types: (1) segmentation-based masks for text-aligned object coverage, (2) random masks for robustness, (3) marginal masks for edge refinement, and (4) interpolation masks for temporal consistency. Each type is applied with a specific probability, and we include a 10\% chance of a null text prompt to encourage general perceptual learning.

This mixed mask strategy exposes the model to diverse spatial and temporal inpainting scenarios, allowing it to adapt within a unified space-time framework. By integrating spatial and temporal tasks in training, the model learns to handle both with improved coherence, as each scenario enhances the other.

% The spatial and temporal inpainting tasks are integrated within the same continuum.To enhance the model’s adaptability across various inpainting scenarios, we employ a mixed mask training strategy that combines four types of masks during training: (1) segmentation-based masks, (2) random masks, (3) marginal masks, and (4) interpolation masks, each applied with a specific probability. Segmentation-based masks, generated using the method above, align with objects or regions referenced by the prompt, supporting precise text-video alignment. Random masks introduce diversity, forcing the model to handle arbitrary masked areas, which improves robustness. Marginal masks, which only cover boundary regions, help the model learn to extend or refine edges, enhancing spatial inpainting capabilities. Finally, interpolation masks focus on masked areas across frames, enabling the model to generate smooth transitions between frames, crucial for temporal consistency.

% For each training instance, we include a 10\% probability of a null text prompt, allowing the model to develop general perceptual and motion skills without relying solely on text guidance. By training with this diverse set of masks, we effectively expose the model to a variety of spatial and temporal inpainting scenarios within a unified space-time framework. Our experiments demonstrate that these different scenarios mutually enhance each other.

%% file: figs/1_1_mask.tex
\begin{figure}
    \centering
    \vspace{-1em}
    \includegraphics[width=\linewidth]{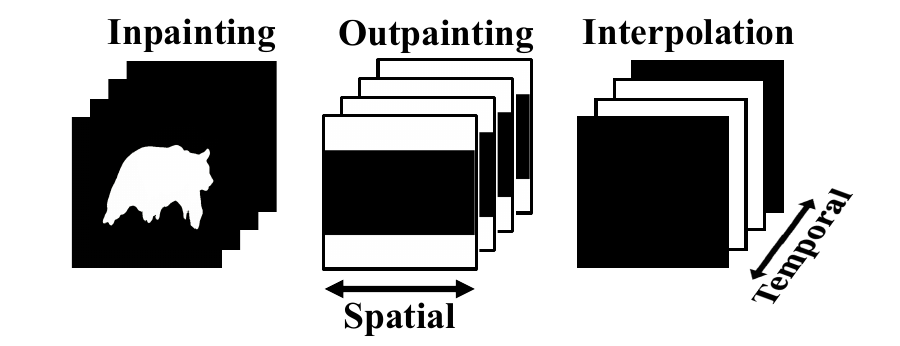}
    \vspace{-1.em}
    \caption{\textbf{Masks for different space-time inpainitng scenarios.} The mask region is shown as white while conserved region is black. In inpainting tasks,
    the masks are continuous regions covering a part of each
    frame. In outpainting tasks, the masks cover the desired
    expansion region. In interpolation tasks, the frames between key frames are masked.
    }
    \label{fig: mask}
    \vspace{-.5em}
\end{figure}

%% file: figs/1_2_mask_generation.tex
\begin{figure}
    \centering
    \vspace{-1em}
    \includegraphics[width=\linewidth]{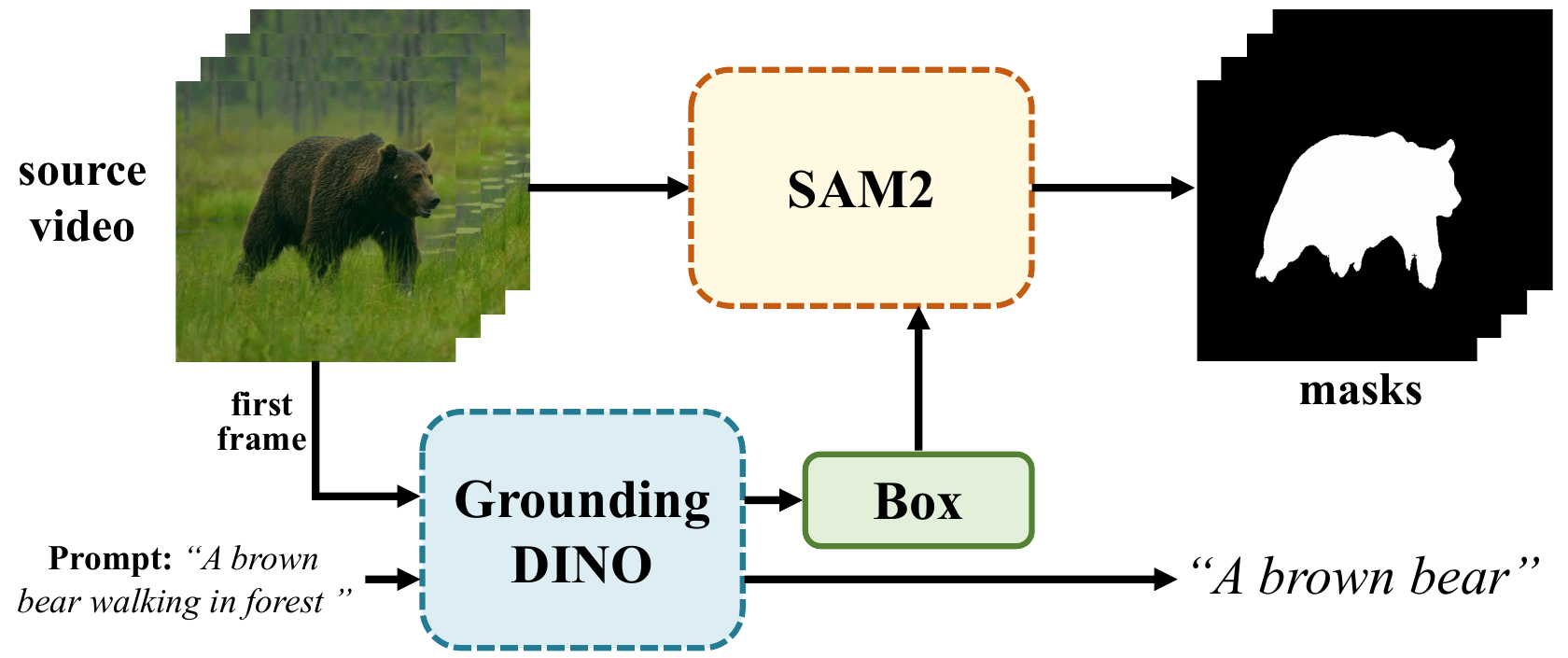}
    \vspace{-1.em}
    \caption{\textbf{Segmentation-based mask generation. } We enhance the training with object-aware masks. We first detect the bounding box and corresponding phrase for objects in the first frame with GroundingDINO\cite{groundingdino_liu2023grounding}, then input the box and source video to SAM2\cite{ravi2024sam2segmentimages}. SAM2 propagates through the video and outputs the corresponding segmentation mask for the grounded object.
    }
    \label{fig: mask_generaiton}
    \vspace{-.5em}
\end{figure}

%% file: sec/4_experiments.tex
\section{Experiemnts}
\input{figs/3_0_qual_comp}
\paragraph{Implementation details.} Our implementation is built upon a StableDiffusion v1.5~\cite{rombach2022high} and AnimateDiff~\cite{animatediff_guo2023animatediff}. Subsequently, the image inpainting layers of the Space-Time Inpainting Adapter are transferred from Brushnet~\cite{ju2024brushnetplugandplayimageinpainting} and frozen during training. For training data, we use the Shutterstock video dataset (Webvid-10M)~\cite{bain2021frozen} and the YoutubeVOS~\cite{xu2018youtube} dataset, with motion modules being trained using 16 frames at a 256×256 resolution with mixed mask selection. We randomly sample from four different types of masks in \cref{sec: mix_mask_training} with probabilities of 0.4, 0.1, 0.2 and 0.3, respectively. For the MoE attention module, we set the number of experts to 4. The motion module and the gating network are trained at the same time, with the rest of the model frozen.
% \paragraph{Training details.} 
During the training stage, the Shutterstock video dataset is watermarked, which would corrupt the model's output if naively trained with. To tackle this problem, we propose a two-stage training procedure. We first train the model on the larger Shutterstock video dataset with $lr=1 \times 10^{-4}$, then finetune the model on the smaller high-quality dataset YoutubeVOS\cite{xu2018youtube} with $lr=1 \times 10^{-5}$. This efficiently alleviated the defects in generation results. 
% In inference stage, 
% \paragraph{Inference details.} 
In the inference stage, we follow DDIM~\cite{song2020denoising}, using 100 sampling steps and the classifier-free guidance scale is 12.5. The mask per frame can be obtained by GroundingDINO~\cite{groundingdino_liu2023grounding}, SAM2~\cite{ravi2024sam2segmentimages} automatically or provided by the users.

\paragraph{Qualitative results}
To comprehensively evaluate the capabilities of our method, we test it on videos across various scenarios, shown in \cref{fig: qual}. Our mask-conditioned inference approach is capable of performing diverse inpainting types, catering to a wide range of mask shapes. Our method adeptly modifies the specified region without affecting the surrounding content and keeps inpainted region consistent with the unmodified area both spatially and temporally.

\subsection{Comparisons}

We present a comprehensive evaluation of our method against other diffusion-based video inpainting techniques, notably VideoComposer~\cite{wang2023videocomposer}, CoCoCo~\cite{zi2024cococoimprovingtextguidedvideo}, AVID\cite{zi2024cococoimprovingtextguidedvideo}, LDMVFI~\cite{Danier_2024} and VIDIM ~\cite{jain2024videointerpolationdiffusionmodels}. The quantitative experiments are conducted on DAVIS dataset~\cite{pont2017davis} containing 200 videos.

\input{tab/quan_spatial}
\input{tab/quan_temporal}

\paragraph{Qualitative comparisons.} \cref{fig: comp: qual} compares the performance on inpainting. Since AVID is not open-source at the time we conduct the experiments, we directly use the cases they picked. The inputs for CoCoCo~\cite{zi2024cococoimprovingtextguidedvideo} and our method are the masks and masked frames, while for VideoComposer~\cite{wang2023videocomposer} we use additional control conditions like structure guidance for better result. Despite more control conditions, VideoComposer shows undesirable generation quality, with watermarks, poor temporal consistency and text alignment. AVID shows poor textual alignment, failing to assign correct colors to the object. CoCoCo fails to capture the overall motion information, painting the car in the wrong direction. Our method not only generates better quality results but shows better temporal consistency and semantic alignment. Please refer to supplementary materials for more qualitative comparisons.

\paragraph{Quantitative comparisons.} Our model’s performance is further quantified using multiple automatic evaluation metrics. For spatial inpainting, we compare background preservation, textual alignment and temporal consistency. Background preservation (BP) is measured using the L1 distance between the original and the edited videos within unaltered regions. The textual alignment (TA) of the generated video is evaluated using the CLIP-score. Temporal consistency (TC) is assessed by computing the cosine similarity between consecutive frames in the CLIP-Image feature space, as per AVID~\cite{zhang2024avidanylengthvideoinpainting} and CoCoCo~\cite{zi2024cococoimprovingtextguidedvideo}. For temporal inpainting, we report the following metrics: peak-signal-to-noise-ratio (PSNR), structural similarity (SSIM)~\cite{wang2004imagessim}, LPIPS\cite{zhang2018unreasonablelpips} and FVD~\cite{unterthiner2018towardsfvd}. As shown in \cref{tab: quan_spatial,tab: quan_temporal}, our model exhibits excellent temporal consistency without compromising per-frame quality. 

\subsection{Ablation analysis}
\paragraph{Space-time inpainting adapter.}
\input{figs/4_0_control}
In \cref{fig: ab_control} we exhibit the effects of varying the mask-conditioned control scale, during the editing of a video of 16 frames. We highlight the first, middle, and last frames to demonstrate how mask-conditioned control impacts the outcomes. For inpainting tasks, a higher control scale ensures background preservation and provide sufficient contextual guidance for the main branch, at the same time restricting the shape of the generated content. This control scale parameter allows users to effectively control the extent of unmasked region protection during the editing process. By manipulating the scale parameter, users can achieve fine-grained control, enabling precise and customizable inpainting.
\paragraph{MoE Attention.}
\input{figs/4_1_moe}
Through mixed masking strategies in our training procedure, we train a MoE attention to actively adapt to different editing cases. With quantitative ablation shown in \cref{tab: quan_spatial,tab: quan_temporal}, we further present a qualitative analysis of MoE attention. In our experiments, we set the number of experts to 4. By replacing the gating network with a deterministic gate, we can set arbitrary weight to the gate and thus look into each expert's capability. \cref{fig: ab_moe} shows the output of each individual expert and the output of MoE attention. We can see that some experts specialize in spatial outpainting while some focus on temporal inpainting. Our gating mechanism is able to smoothly adapt to each case by reading the shape of the mask input.

%% file: figs/3_0_qual_comp.tex
\begin{figure*}
\setlength{\linewidth}{\textwidth}
\setlength{\hsize}{\textwidth}
\centering
\includegraphics[width=1\linewidth]{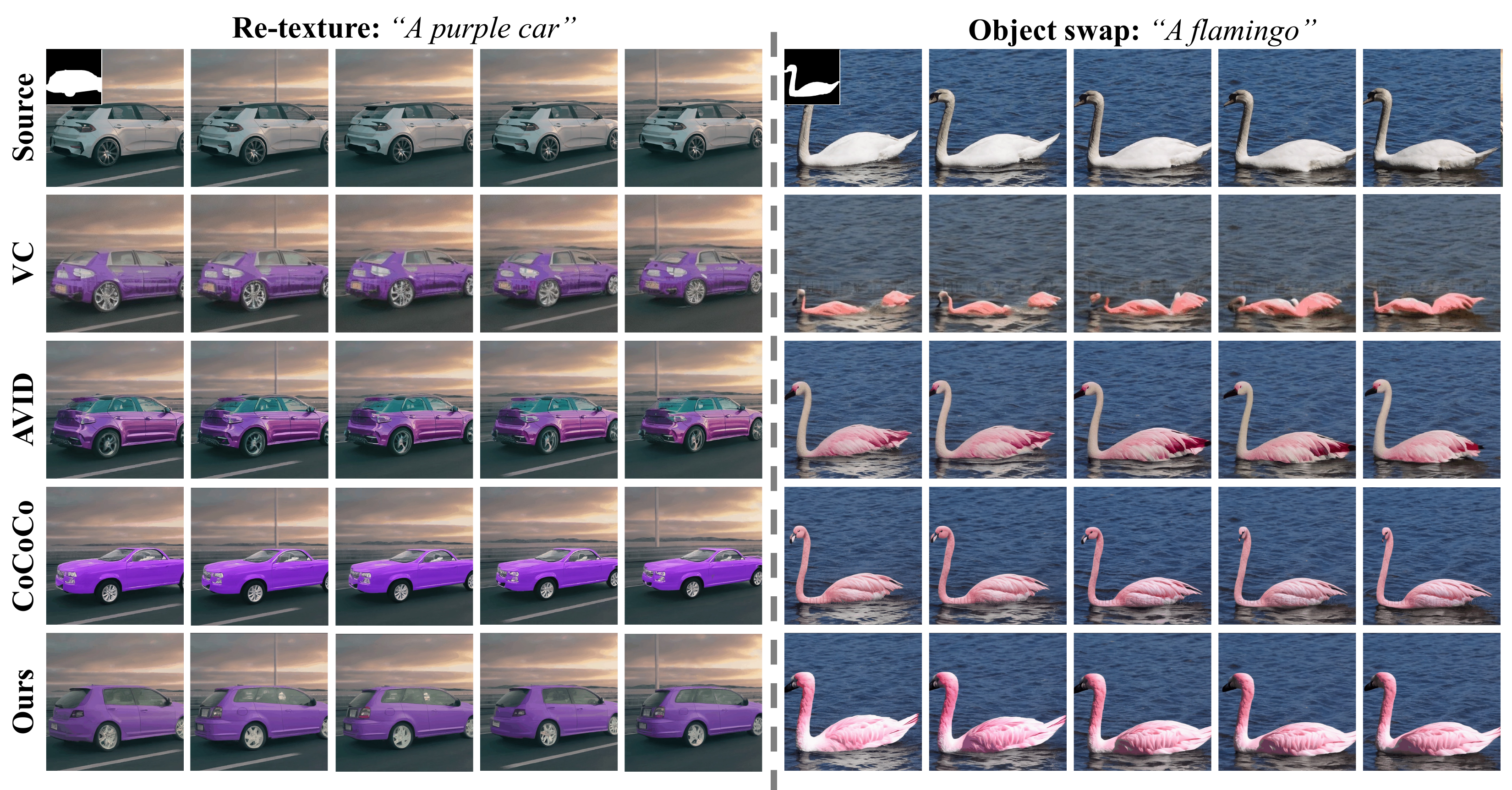}
\caption{\textbf{Comparison with previsou methods.} 
We compare our method against several approaches, including VideoComposer\cite{wang2023videocomposer}, AVID\cite{zhang2024avidanylengthvideoinpainting} and CoCoCo\cite{zi2024cococoimprovingtextguidedvideo}. The results of AVID directly comes from its puclication, other methods are evaluated using their default hyper-parameters as specified in source codes. Each video in our experiments consists of 16 frames. Our method successfully inpaints the masked region following text prompt with remarkable consistency. Notably, our method demonstrate better texual alignment and visual quality than other methods in our comparison.
}
\label{fig: comp: qual}

\vspace{-1em}
\end{figure*}

%% file: tab/quan_spatial.tex
\begingroup
\setlength{\tabcolsep}{3pt} %
\renewcommand{\arraystretch}{0.9} %
\vspace{-.5em}
\begin{table}\centering
\small
\begin{tabular*}{1.0\linewidth}{@{\extracolsep{\fill}}lccc|ccc}

Task & \multicolumn{3}{c}{Inpainting} & \multicolumn{3}{c}{Outpainting} \\
\toprule 
Metric & BP$\downarrow$ & TA$\uparrow$ & TC$\uparrow$ & BP$\downarrow$ & TA$\uparrow$ & TC$\uparrow$\\
\midrule
VC~\cite{wang2023videocomposer}     & 47.9 & 31.0 & 96.4 & 46.7 & 31.2 & 96.2  \\
CoCoCo~\cite{zi2024cococoimprovingtextguidedvideo}  & 42.3 & 31.4 & \textbf{97.6} & 42.1 & 31.3 & 97.0  \\
Ours Spatial.    & 42.2 & 31.4 & 97.1 & 41.9 & 31.2 & 97.1\\
Ours w/o MoE    & 42.3 & 31.2 & 97.3 & 42.0 & 31.3 & 96.9\\
Ours    & \textbf{41.8} & \textbf{31.8} & 97.5 & \textbf{41.5} & \textbf{31.6} & \textbf{97.4}\\
\bottomrule 
\end{tabular*}
\caption{\textbf{Quantitative results on spatial inpainting.} We compare our method against several spatial inpainting model, including s VideoComposer~\cite{wang2023videocomposer}, CoCoCo~\cite{zi2024cococoimprovingtextguidedvideo}. Ours Spatial. is trained only with spatial inpainting cases. Ours w/o MoE repalces MoE with single FFN. BP, TA, TC represent background preservation, textual alignment, temporal consistency, respectively. The best results are marked in \textbf{bold}.
}

\vspace{-.5em}
\label{tab: quan_spatial}
\end{table}
\endgroup

%% file: tab/quan_temporal.tex
\begingroup
\setlength{\tabcolsep}{3pt} %
\renewcommand{\arraystretch}{0.9} %
\vspace{-.5em}
\begin{table}[!htb]\centering
\small
\begin{tabular*}{1.0\linewidth}{@{\extracolsep{\fill}}lcccc}
Task & \multicolumn{4}{c}{Temporal inpainting} \\
\toprule % Top horizontal line
Metric                                                 & PSNR$\uparrow$ & SSIM~\cite{wang2004imagessim}$\uparrow$ & LPIPS~\cite{zhang2018unreasonablelpips}$\downarrow$ & FVD~\cite{unterthiner2018towardsfvd}$\downarrow$\\
\midrule % Middle horizontal line
LDMVFI\cite{Danier_2024}                               & 19.98 & 0.4794 & 0.2764 & 245.02  \\
VIDIM\cite{jain2024videointerpolationdiffusionmodels}  & 19.62 & 0.4709 & 0.2578 & \textbf{199.32} \\
Ours Temporal.                                         & 19.82 & 0.4783 & 0.2599 & 204.53\\
Ours w/o MoE                                           & 19.79 & 0.4769 & 0.2612 & 211.28\\
Ours                                                   & \textbf{20.01} & \textbf{0.4814} & \textbf{0.2547} & 201.35 \\
\bottomrule % Bottom horizontal line
\end{tabular*}
\caption{\textbf{Quantitative results on temporal inpainting.} We compare our method against several diffusion-based temporal inpainting models, including  LDMVFI\cite{Danier_2024} , VIDIM\cite{jain2024videointerpolationdiffusionmodels}. Ours Temporal. is trained only with temporal inpainting cases. PSNR denotes peak-signal-noise-ratio.
}

\vspace{-.5em}
\label{tab: quan_temporal}
\end{table}
\endgroup

%% file: figs/4_0_control.tex
\begin{figure}
    \centering
    \vspace{-1em}
    \includegraphics[width=\linewidth]{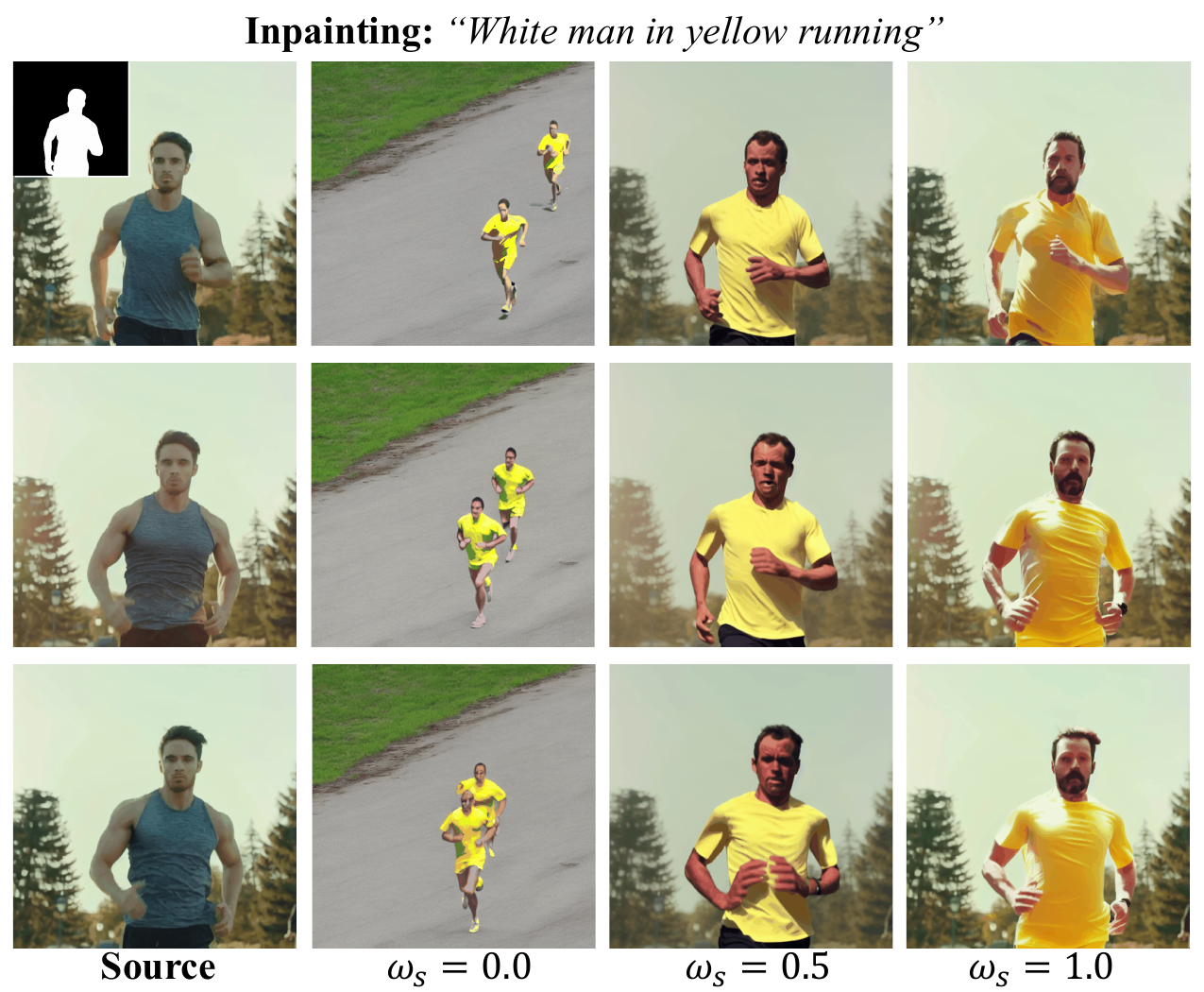}
    \vspace{-1.em}
    \caption{\textbf{Analysis of adapter control scale.} The source video is shown on the left. We show the results of inpainting with control scale 0.0, 0.5 and 1.0. A higher control scale ensures background preservation and contextual guidance.
    } 
    \label{fig: ab_control}
    \vspace{-.5em}
\end{figure}

%% file: figs/4_1_moe.tex
\begin{figure}
    \centering
    \vspace{-.5em}
    \includegraphics[width=\linewidth]{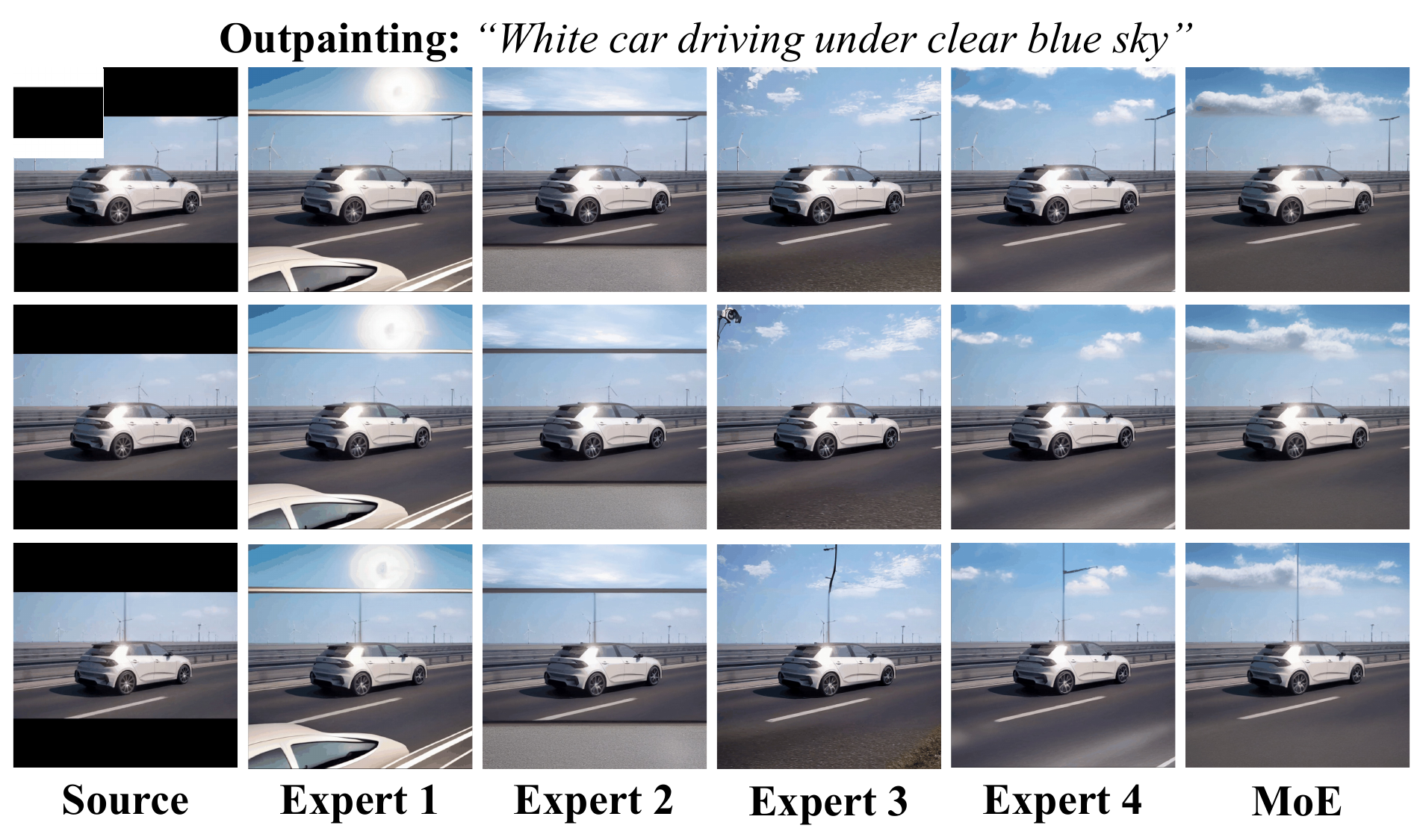}
    
    \caption{\textbf{Analysis of MoE attention.} The source video is shown on the left. We show the results of each individual expert FFN in comparison with the result of MoE. Some experts are more specialized in outpainting tasks.The Mask-Gated MoE adpaptively synthesizes the outputs of experts, showing best consistency.
    }
    \vspace{-.5em}
    \label{fig: ab_moe}
    
\end{figure}

%% file: sec/5_conclusion.tex
\section{Limitation and future work}
Although UniPaint has achieved great space-time video inpainting performance, it still faces challenges when inpainting the source video with large motion. The visual cases can be found in supplementary materials. We analyze that it may be due to the motion bias in the training dataset. We will finetune our approach in larger video dataset. Additionally, in the future, we are considering integrating more tasks into our framework, including video super-resolution(spatial regional mask) and video prediction (temporal outpainting mask).

\section{Conclusion}
\label{sec: conclusion}
In this paper, we present UniPaint, a unified generative space-time video inpainting framework that enables
spatial-temporal inpainting and interpolation. Different
from existing methods that treat video inpainting and video
interpolation as two distinct tasks, we leverage a unified inpainting framework to address them.
To adapt to different text-to-video models, we first introduce a plug-and-play space-time
video inpainting adapter. To cover various tasks, we design the Mixture of Experts (MoE) attention and spatial-temporal masking strategy during the training stage. Our method produces high-quality and aesthetically pleasing results, achieving the best quantitative results across various tasks. We hope our work can pave the way for further progress in this promising direction and push this frontier.

%% file: sec/X_supp.tex
\clearpage
\appendix
\setcounter{section}{0}
\setcounter{figure}{0}
\setcounter{table}{0}
\renewcommand{\thefigure}{A\arabic{figure}}
\renewcommand{\thetable}{A\arabic{table}}

\input{figs/supp_qual}
\section*{Overview}
This supplementary material provides additional details and insights to further elaborate on various aspects of the proposed method. The content is organized as follows:
\begin{itemize}
    \item \textbf{Experiment Details:} Detailed information about the training and evaluation procedures can be found in \cref{sec:exp}.
    \item \textbf{More Qualitative Results:} In \cref{sec:qual_result}, we showcase an expanded set of qualitative experiments, highlighting the flexibility and consistency of our approach.
    \item \textbf{More Comparative Analysis:} Beyond the qualitative comparisons presented in the main paper, \cref{sec:qual_comp} includes further analyses focusing on marginal and temporal inpainting tasks.
    \item \textbf{Limitations:} In \cref{sec:lim}, we discuss the limitations of our method and outline potential areas for future improvement.

\end{itemize}
\input{figs/supp_out}
% \section{Experiment Details}
% \label{sec:exp}
% Our model is trained with a two-stage procedure. We first train our model on Webvid-10M~\cite{webvid_bain2021frozen} for 5 epochs with $lr=1 \times 10^{-4}$, then train on YoutubeVOS~\cite{xu2018youtube} dataset for 10 epochs with $lr=1 \times 10^{-5}$. We use AdamW~\cite{adamw_loshchilov2017decoupled} as optimizer. The training process is conducted on 8 Nvidia A100 GPUs for around 3 days. All ablations are trained using the same settings. For inference, UniPaint requires 30 GB memory with float16 precision, the inference process for one video clip takes 69 seconds on one Nvidia A100 GPU.

\section{Experiment Details}
\label{sec:exp}
Our model is trained using a two-stage procedure:
\begin{enumerate}
    \item \textbf{Initial Training:} The model is first trained on the WebVid-10M dataset~\cite{webvid_bain2021frozen} for 5 epochs using a learning rate of \(1 \times 10^{-4}\).
    \item \textbf{Fine-Tuning:} Fine-tuning is conducted on the YouTubeVOS dataset~\cite{xu2018youtube} for 10 epochs with a reduced learning rate of \(1 \times 10^{-5}\).
\end{enumerate}

We utilize the AdamW optimizer~\cite{adamw_loshchilov2017decoupled} for both stages of training. The process is performed on 8 NVIDIA A100 GPUs over approximately 3 days. All ablation studies follow the same training configuration for consistency.

For inference, UniPaint operates in float16 precision and requires 30 GB of GPU memory. Processing a single video clip takes 69 seconds on one NVIDIA A100 GPU.

\section{Qualitative Results}
\label{sec:qual_result}
As illustrated in \cref{fig: supp_qual}, we present additional qualitative results demonstrating the capabilities of our method across diverse inpainting scenarios. \emph{UniPaint} exhibits strong adaptability and maintains consistent performance across a variety of inpainting tasks.

\paragraph{Object Removal.}
Early works on video inpainting often focused on object removal as a primary task~\cite{xu2019deep,kim2019deep,zeng2020learning}. While modern diffusion models provide greater generative flexibility, our method retains the ability to perform effective object removal. By applying appropriate masks and providing textual prompts that describe the desired background, \emph{UniPaint} can efficiently eliminate unwanted objects from video sequences while maintaining spatial and temporal consistency.

\paragraph{Environment Swap.}
Environment swapping can be considered a specialized case of outpainting. By selecting the complement of the target region as the editing area, our method enables seamless integration of a foreground object into a custom background. Using prompts that describe the new environment, \emph{UniPaint} accurately modifies the scene, ensuring that the object appears naturally within the specified setting.

\section{Quanlitative Comparisons}
\label{sec:qual_comp}
\input{figs/supp_interpolation}

We further conduct more comparative analysis against various inpainting models, as shown in ~\cref{fig: supp_out,fig: supp_interpolation}. 

\paragraph{Outpainting.}For outpainting, we compare our method with AVID~\cite{zhang2024avidanylengthvideoinpainting}, CoCoCo~\cite{zi2024cococoimprovingtextguidedvideo}. As shown in ~\cref{fig: supp_out}, our method significantly outperforms both AVID and CoCoCo. AVID exhibits noticeable artifacts and blending issues in the outpainted regions, failing to maintain texture and scene consistency. CoCoCo produces more coherent outputs than AVID but lacks fine-grained alignment with the original scene, resulting in less natural extensions. In contrast, our method generates sharp, realistic, and seamlessly integrated outpainting results, preserving both structural and textural fidelity to the original scene.

\paragraph{Temporal inpainting.}For temporal inpainting, we compare our method with RIFE~\cite{huang2022rife} and LDMVFI~\cite{Danier_2024}. As shown in ~\cref{fig: supp_interpolation}, our method achieves the best interpolation quality among the evaluated models. RIFE outputs suffer from blurriness and inconsistent details, particularly at object edges and in motion dynamics. LDMVFI demonstrates better temporal coherence than RIFE but introduces blending artifacts and lacks sharpness in reconstructed details, like the wheels of the bus. Our approach produces the most consistent and realistic temporal inpainting, maintaining fine details, sharp edges, and seamless transitions across frames, ensuring both visual fidelity and temporal smoothness.

\section{Limitations}
\label{sec:lim}
\input{figs/supp_fail}

Despite the promising results achieved by our proposed method, several limitations remain, particularly in handling complex and dynamic scenes. As shown in the failure cases in ~\cref{fig: supp_fail}, our model struggles to maintain accurate body proportions and motion coherence when dealing with intricate human movements, such as breakdancing or snowboard tricks. Artifacts such as unnatural poses, distorted body parts, and inconsistent blending are common in these scenarios, suggesting that further advancements in motion understanding and temporal consistency are required.

While our method performs well on common inpainting tasks, we admit that it struggles with rare or unseen scenarios, such as unconventional poses or extreme actions. This limitation most possibly stems from the training data, which may not comprehensively cover all possible variations in motion and context. Addressing these limitations is an essential direction for future work. Incorporating advanced motion priors, leveraging larger and more diverse datasets, and optimizing the model’s efficiency will improve its robustness and applicability to real-world video generation tasks.

%% file: figs/supp_qual.tex
\twocolumn[{
\renewcommand\twocolumn[1][]{#1}
\maketitlesupplementary
\begin{center}
    \centering
    \vspace*{-.5em}
    \includegraphics[width=1\linewidth]{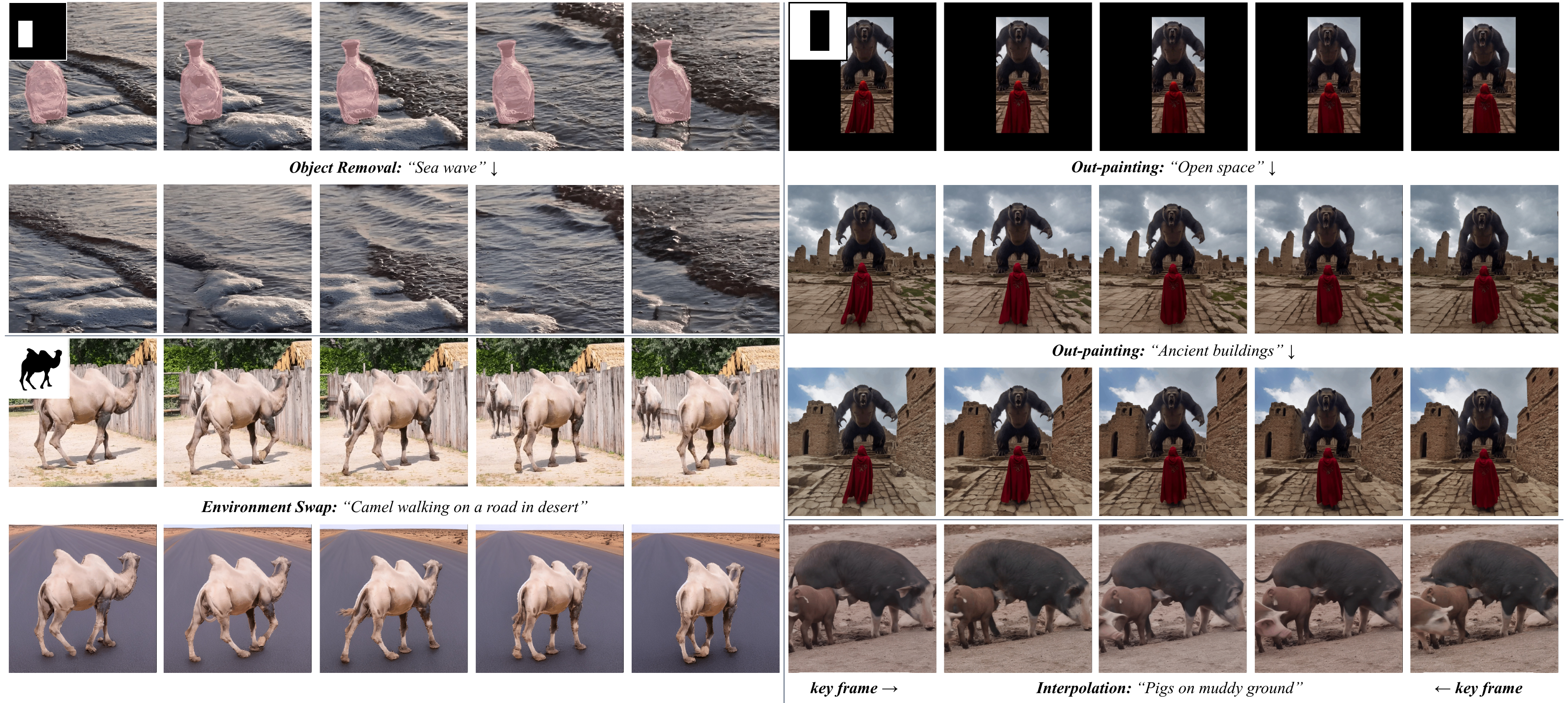}
    \captionof{figure}{
    \textbf{More results of our method.} Additional qualitative results of our method applied to various inpainting scenarios, including object removal, environment swapping, outpainting, and temporal inpainting(interpolation). These examples demonstrate the flexibility of our method across diverse scenarios.
    }
    \label{fig: supp_qual}
\end{center}
}]

%% file: figs/supp_out.tex
\begin{figure}
    \centering
    \vspace{-1em}
    \includegraphics[width=\linewidth]{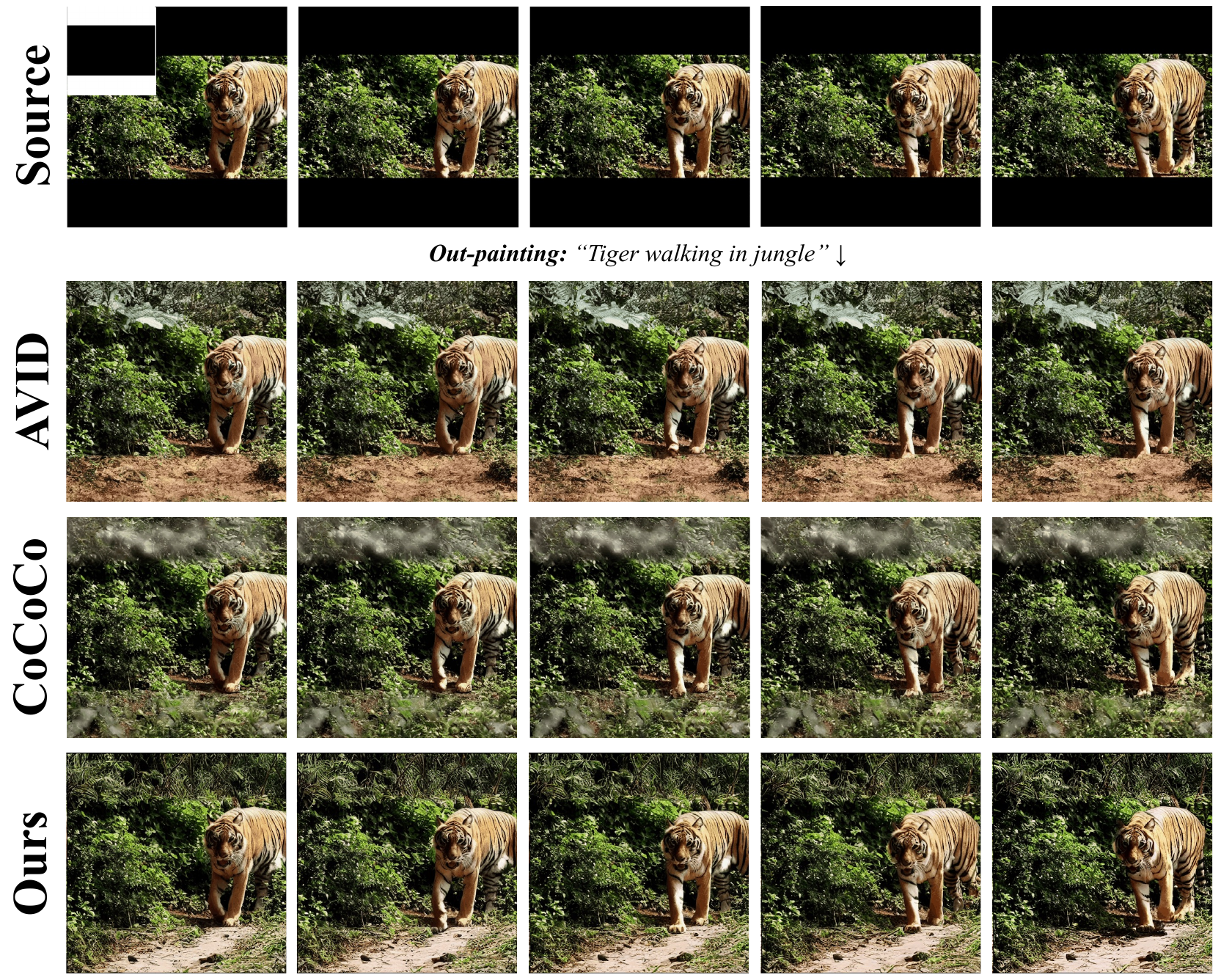}
    \vspace{-1.em}
    \caption{\textbf{Comparative analysis of space-time video outpainting.} We compare our method against AVID~\cite{zhang2024avidanylengthvideoinpainting} and CoCoCo~\cite{zi2024cococoimprovingtextguidedvideo} for outpainting. Our method achieves the most realistic and coherent outpainting, with superior alignment, texture consistency, and scene continuity compared to AVID and CoCoCo.
    }
    \label{fig: supp_out}
    \vspace{-.5em}
\end{figure}

%% file: figs/supp_interpolation.tex
\begin{figure}
    \centering
    \vspace{-1em}
    \includegraphics[width=\linewidth]{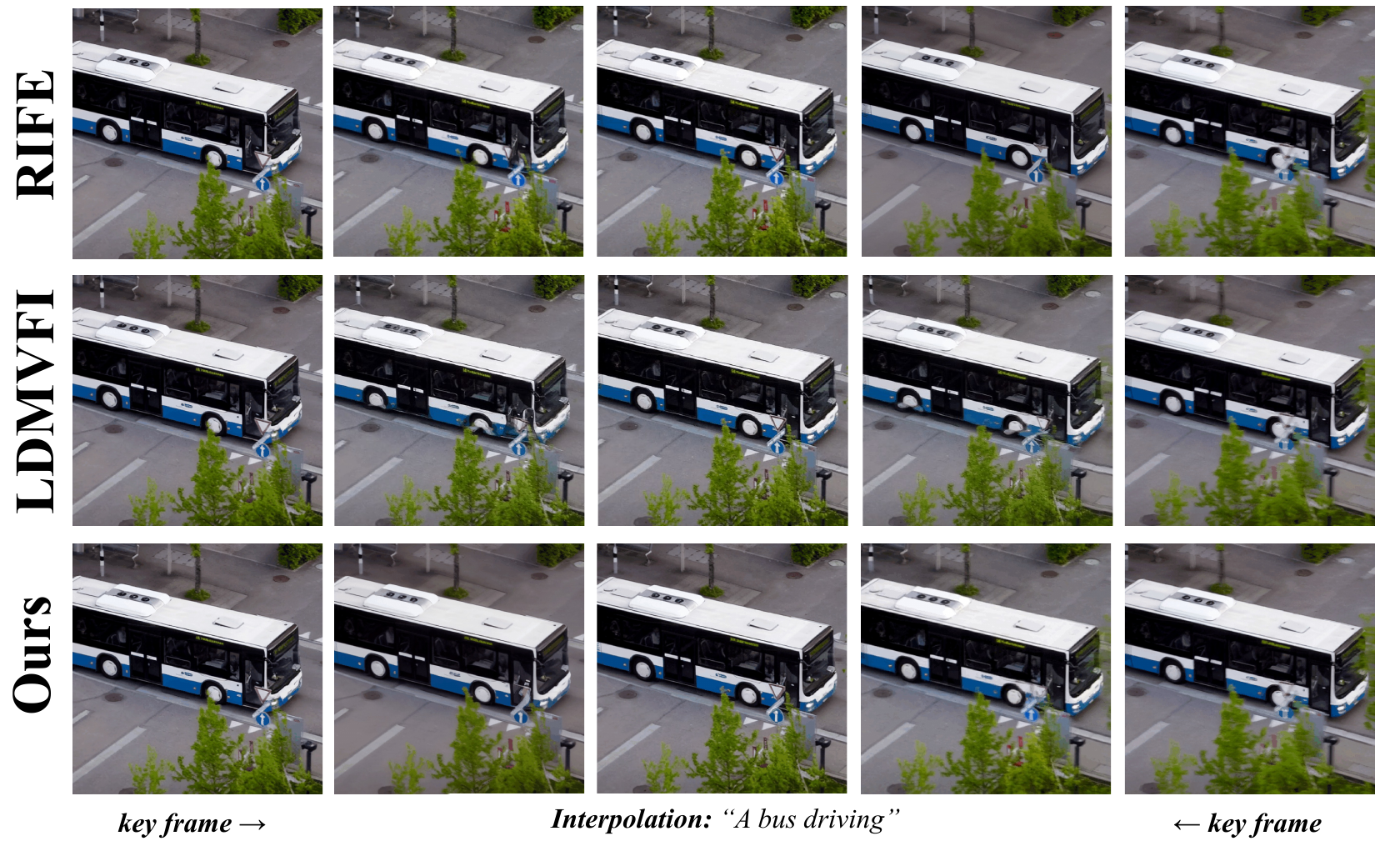}
    \vspace{-1.em}
    \caption{\textbf{Comparative analysis on temporal inpainting.} Key frames are provided at the beginning and end, with interpolated frames shown for RIFE~\cite{huang2022rife}, LDMVFI~\cite{Danier_2024}, and our method. RIFE shows blurriness and inconsistent details, while LDMVFI exhibits better temporal coherence but introduces blending artifacts and lacks sharpness. \emph{UniPaint} achieves the most realistic and consistent temporal inpainting, preserving fine details, sharp edges, and seamless transitions across all frames.
    }
    \label{fig: supp_interpolation}
    \vspace{-.5em}
\end{figure}

%% file: figs/supp_fail.tex
\begin{figure}
    \centering
    \vspace{-1em}
    \includegraphics[width=\linewidth]{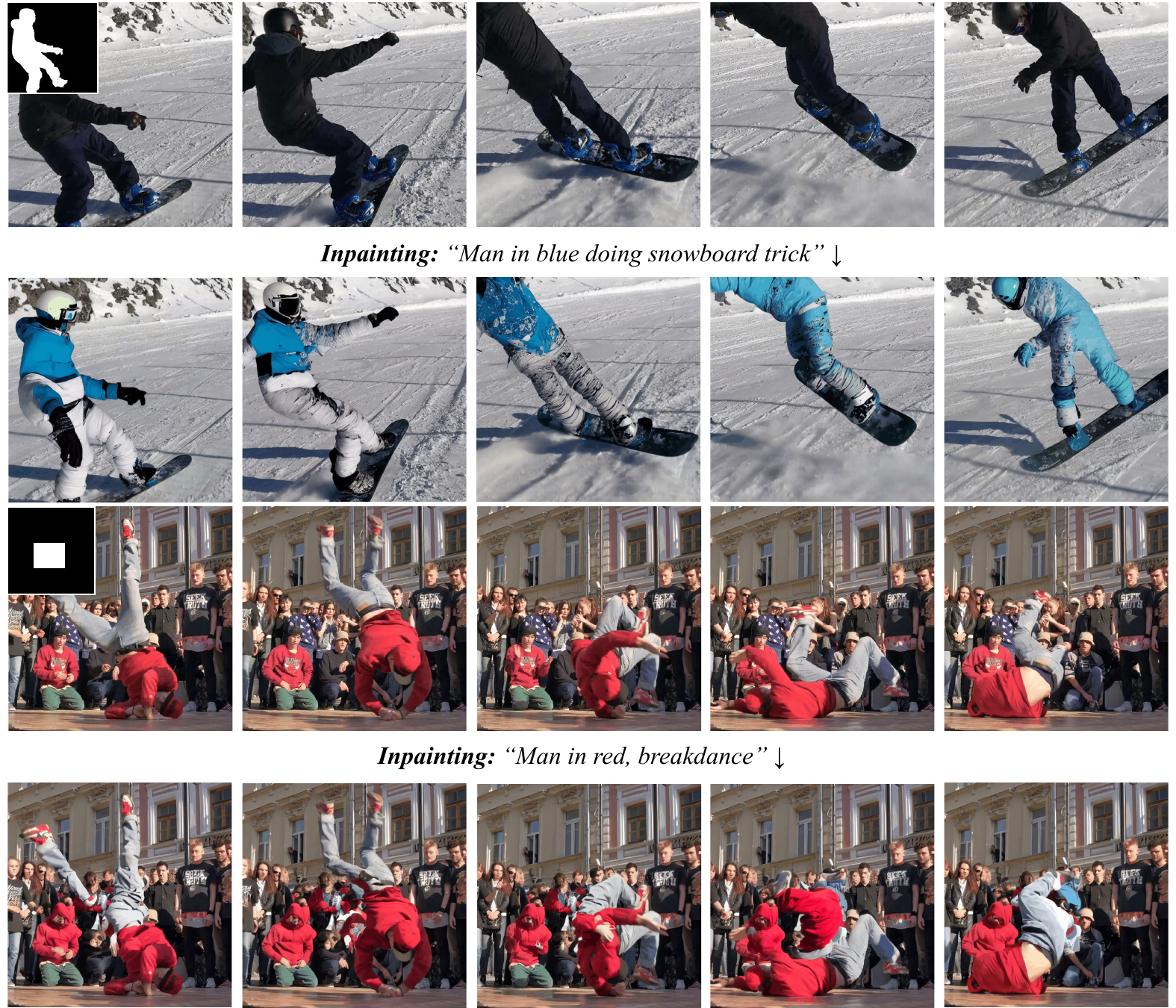}
    \vspace{-1.em}
    \caption{\textbf{Failure cases.} Our method fails to generate results with complex movements. For the snowboard trick (top), the generated sequence struggles with body proportions, motion dynamics, and texture blending. In the breakdance case (bottom), the dancer’s movements and interactions with the environment lack coherence, and the surrounding crowd suffers from visual artifacts.
    }
    \label{fig: supp_fail}
    \vspace{-.5em}
\end{figure}